\documentclass[11pt]{article}

\usepackage[utf8]{inputenc}
\usepackage[T1]{fontenc}
\usepackage{lmodern}
\usepackage{amsmath,amssymb,amsfonts}
\usepackage{booktabs}
\usepackage{caption}
\usepackage{subcaption}
\usepackage{hyperref}
\hypersetup{
  pdftitle={Hydra: Unifying Document Retrieval and Generation in a Single Vision-Language Model},
  pdfauthor={Athos Georgiou}
}
\usepackage{cleveref}
\usepackage{xcolor}
\usepackage{algorithm}
\usepackage{algorithmic}
\usepackage{enumitem}
\usepackage{needspace}
\usepackage{float}
\usepackage[margin=1in]{geometry}
\usepackage{natbib}
\usepackage{url}
\usepackage{pgfplots}
\pgfplotsset{compat=1.18}
\usetikzlibrary{positioning,arrows.meta,calc,patterns,decorations.pathreplacing}

\newcommand{\ours}{Hydra}

\newcommand{\ndcg}{nDCG@5}

\title{\ours{}: Unifying Document Retrieval and Generation\\in a Single Vision-Language Model}

\author{
  Athos Georgiou \\
  Independent Researcher \\
  \texttt{athos.georgiou@nca-it.com}
}

\date{April 2026}

\begin{document}
\maketitle

\begin{abstract}
Visual document understanding typically requires separate retrieval and generation models, doubling memory and system complexity.
We present \ours{}, a dual-head approach that provides both ColBERT-style late-interaction retrieval and autoregressive generation from a single vision-language model.
A single LoRA adapter, trained only for retrieval, is toggled at inference: enabling it produces multi-vector embeddings; disabling it recovers the base model's generation quality, with 426 of 426 language-model weight tensors byte-for-byte identical to a freshly-loaded Qwen3.5-4B.
We identify two failure modes that can silently break generation in retrieval-fine-tuned VLMs (attention-mode restoration and \texttt{lm\_head} preservation) plus an efficiency requirement (KV-cache-aware decoding); \ours{} sidesteps the first two structurally and addresses the third in the decode loop.
We release two scales, \textbf{Hydra-4B} and \textbf{Hydra-0.8B}, sharing LoRA hyperparameters ($r{=}32$, $\alpha{=}32$) and optimisation recipe; data mix and projection dim differ across scales.
The single-model design cuts peak GPU memory from 28.85\,GB to 10.77\,GB at 4B (62.7\% reduction) and from 5.79\,GB to 2.37\,GB at 0.8B (59.1\%) relative to a co-resident two-model deployment.
A controlled ablation finds GritLM-style joint training matches \ours{}'s retrieval-only training on the evaluated modes while its LoRA-on generation mode collapses.
A proof-of-concept on Qwen2.5-Omni-3B preserves generation equivalence on a non-Qwen3.5 backbone and transfers image retrieval within 2--8\,pp of \ours{}-4B, with zero-shot audio retrieval emerging through the frozen Whisper encoder.

\end{abstract}

\section{Introduction}
\label{sec:intro}
Document AI systems must solve two distinct tasks: \emph{retrieval} (finding relevant pages given a query) and \emph{understanding} (extracting and interpreting information within those pages).
Modern approaches address these with separate models: a retrieval model such as ColPali~\citep{faysse2025colpali} or ColQwen2~\citep{colqwen2} for page-level retrieval via ColBERT-style late interaction~\citep{khattab2020colbert,santhanam2022colbertv2}, and a generative VLM such as Qwen2.5-VL~\citep{bai2025qwen25vl} for document understanding.
This dual-model paradigm is wasteful.
Both models share a common backbone architecture (a vision-language transformer), yet they must be loaded independently, doubling GPU memory requirements and complicating deployment.

The waste is particularly stark: ColPali-family models \emph{are} fine-tuned VLMs.
ColPali, ColQwen2, and ColQwen3.5~\citep{colqwen35} all begin from a pretrained VLM, add a linear projection head (\texttt{custom\_text\_proj}) for 128- or 320-dimensional multi-vector embeddings, and fine-tune with contrastive loss.
This fine-tuning modifies the model's attention patterns (to bidirectional) and internal representations, sacrificing autoregressive generation, though the capability remains latent beneath the retrieval-adapted weights.

We observe that this sacrifice is unnecessary when using Low-Rank Adaptation (LoRA)~\citep{hu2022lora}.
Because LoRA adapters are additive ($W_{\text{adapted}} = W_{\text{base}} + BA$), disabling them at inference time \emph{exactly} recovers the base model's weights.
This means a single VLM with a retrieval LoRA adapter can serve as both:
\begin{itemize}[nosep]
    \item A \textbf{retrieval model} (LoRA-on, bidirectional attention $\rightarrow$ \texttt{custom\_text\_proj} $\rightarrow$ 320-dim embeddings), and
    \item A \textbf{generative VLM} (LoRA-off, causal attention $\rightarrow$ \texttt{lm\_head} $\rightarrow$ autoregressive text).
\end{itemize}
Critically, only the retrieval head requires training.
Generation capability is recovered by disabling the adapter and restoring causal attention, though realizing this in practice requires addressing three non-obvious engineering requirements (\Cref{sec:requirements}).
Prior work has explored related ideas.
SV-RAG~\citep{chen2025svrag} trains \emph{two} LoRA adapters on a shared VLM (one for retrieval, one for generation) and swaps them at inference.
URaG~\citep{shi2026urag} unifies both tasks by inserting a retrieval module at an intermediate transformer layer.
ColQwen2\_4RAG~\citep{oprea2025colqwen24rag} demonstrated the same LoRA on/off toggling mechanism in an application setting; we add a systematic analysis of failure modes that \emph{can} occur under different training setups (\Cref{sec:requirements}), a controlled-baseline retrieval evaluation, and a comparison against joint training.
GritLM~\citep{muennighoff2025gritlm} showed that joint training can unify embedding and generation in text-only models.
Our contribution is not the toggling mechanism itself, which exists in prior work, but a systematic characterisation of when and why it works: identifying two failure modes that silently break generation without these safeguards, plus an efficiency requirement, and demonstrating through controlled experiments that generation training is unnecessary at the LoRA ranks tested.

We call this architecture \ours{}: one model, two heads.\footnote{We release two instantiations on the Qwen3.5 family: \textbf{HydraQwen3.5-4B} (4.60B parameters, retrieval dim 320) and \textbf{HydraQwen3.5-0.8B} ($\approx$800M parameters, retrieval dim 128). Full training configuration is in \Cref{sec:training}.}
\Cref{fig:architecture} illustrates the architecture, and \Cref{fig:rag-pipeline} shows how this extends to a complete retrieval-augmented generation (RAG) pipeline.
Our contributions are:

\begin{enumerate}[nosep]
    \item A \textbf{dual-head approach} that provides both ColBERT retrieval and autoregressive generation from a single VLM, requiring only a single LoRA adapter and no generation training. We characterise the operating constraints for making this work: two correctness requirements (attention-mode restoration, \texttt{lm\_head} preservation) that \ours{} sidesteps structurally, plus a practical efficiency requirement (KV-cache support) addressed in the decode loop (\Cref{sec:method}).

    \item \textbf{Two-scale evaluation} of the mechanism: retrieval on 22 ViDoRe tasks (V1/V2/V3) at both 4B and 0.8B scales, a bitwise generation-equivalence audit showing 426/426 language-model weight tensors match a freshly-loaded base model after LoRA is disabled, and efficiency measurements demonstrating that the VRAM-savings ratio is stable across scale (\Cref{sec:experiments}).

    \item A \textbf{controlled ablation} at $r{=}16$ comparing \ours{}'s retrieval-only training against GritLM-style joint training; the two are equivalent on retrieval and LoRA-off generation, while the LoRA-on generation mode that joint training was designed to enable collapses (\Cref{sec:ablation}).
\end{enumerate}

\begin{figure*}[t]
\centering
\definecolor{steelblue}{RGB}{55,126,184}
\definecolor{warmred}{RGB}{200,80,80}
\begin{tikzpicture}[
    box/.style={rounded corners=3pt, minimum height=0.9cm, minimum width=2.2cm,
                align=center, font=\sffamily\small, line width=0.7pt},
    smallbox/.style={rounded corners=2pt, minimum height=0.7cm, minimum width=1.8cm,
                     align=center, font=\sffamily\footnotesize, line width=0.7pt},
    arrow/.style={-{Stealth[length=6pt, width=5pt]}, line width=0.9pt},
    darrow/.style={-{Stealth[length=6pt, width=5pt]}, line width=0.9pt, dashed},
    lbl/.style={font=\sffamily\footnotesize\itshape, text=gray!60!black},
]

\node[box, draw=gray!60, fill=gray!8, minimum width=3.5cm]
    (vlm) at (0,0) {Qwen3.5\\(frozen $W_{\text{base}}$)};

\node[box, draw=gray!60, fill=gray!8, minimum width=3.5cm]
    (vision) at (0,-1.8) {Vision Encoder\\(frozen)};

\node[box, draw=gray!70, fill=gray!5, minimum width=2.8cm, line width=0.9pt]
    (lora) at (0,2.0) {LoRA Adapter\\($r{=}32$, $\alpha{=}32$)};

\node[smallbox, draw=steelblue, fill=steelblue!15]
    (bidir) at (-5.0,2.0) {Bidirectional\\Attention};
\node[smallbox, draw=steelblue, fill=steelblue!15, minimum width=2.2cm]
    (proj) at (-5.0,0) {\texttt{custom\_text\_proj}\\$\mathbb{R}^d \to \mathbb{R}^{320}$};
\node[box, draw=steelblue, fill=steelblue!20, minimum width=2.8cm]
    (emb) at (-5.0,-1.8) {320-dim Embeddings\\(MaxSim scoring)};

\node[smallbox, draw=warmred, fill=warmred!12]
    (causal) at (5.0,2.0) {Causal\\Attention};
\node[smallbox, draw=warmred, fill=warmred!12, minimum width=2.2cm]
    (lmhead) at (5.0,0) {\texttt{lm\_head}\\$\mathbb{R}^d \to \mathbb{R}^{|V|}$};
\node[box, draw=warmred, fill=warmred!15, minimum width=2.8cm]
    (gen) at (5.0,-1.8) {Autoregressive Text\\(KV-cache)};

\node[font=\sffamily\small\bfseries, text=steelblue]
    at (-5.0, 3.1) {Retrieval Mode};
\node[lbl] at (-5.0, 2.7) {LoRA-on};
\node[font=\sffamily\small\bfseries, text=warmred]
    at (5.0, 3.1) {Generation Mode};
\node[lbl] at (5.0, 2.7) {LoRA-off};

\draw[arrow, steelblue] (bidir.south) -- (proj.north);
\draw[arrow, steelblue] (proj.south) -- (emb.north);

\draw[arrow, warmred] (causal.south) -- (lmhead.north);
\draw[arrow, warmred] (lmhead.south) -- (gen.north);

\draw[arrow, gray!60] (vision.north) -- (vlm.south);

\draw[arrow] (vlm.north) -- (lora.south);
\draw[arrow, steelblue] (lora.west) -- (bidir.east);
\draw[darrow, warmred] (lora.east) -- (causal.west);

\end{tikzpicture}
\caption{%
\ours{} architecture. A single VLM serves two modes by toggling a LoRA adapter at inference time.
\textbf{Left}: Retrieval mode (LoRA-on, bidirectional attention) produces multi-vector embeddings via \texttt{custom\_text\_proj} (320-dim shown here for \ours{}-4B; \ours{}-0.8B uses 128-dim).
\textbf{Right}: Generation mode (LoRA-off, causal attention) produces autoregressive text via the base \texttt{lm\_head} with KV-cache.
The vision encoder is frozen and shared. No weight copying or model reloading occurs between modes.
Arrow style indicates the data path active in each mode: solid blue = LoRA-on retrieval flow; dashed red = LoRA-bypassed generation flow (the dashed line marks that LoRA is disabled along this path, not that the layer is absent).
}
\label{fig:architecture}
\end{figure*}

\begin{figure*}[t]
\centering
\definecolor{steelblue}{RGB}{55,126,184}
\definecolor{warmred}{RGB}{200,80,80}
\resizebox{\textwidth}{!}{%
\begin{tikzpicture}[
    box/.style={rounded corners=3pt, minimum height=0.85cm,
                align=center, font=\sffamily\small, line width=0.7pt},
    modelbox/.style={box, draw=steelblue, line width=0.9pt,
                     fill=steelblue!12},
    genbox/.style={box, draw=warmred, line width=0.9pt,
                   fill=warmred!12},
    databox/.style={box, draw=gray!60, fill=gray!8,
                    minimum width=1.6cm, font=\sffamily\footnotesize},
    hydrabox/.style={box, draw=gray!70, fill=gray!5,
                     minimum height=1.0cm, line width=0.9pt},
    headlabel/.style={rounded corners=2pt, minimum width=1.6cm,
                      font=\sffamily\scriptsize, minimum height=0.45cm,
                      align=center, line width=0.7pt},
    retlabel/.style={headlabel, draw=steelblue, fill=steelblue!18},
    genlabel/.style={headlabel, draw=warmred, fill=warmred!18},
    arrow/.style={-{Stealth[length=6pt, width=5pt]}, line width=0.9pt},
    idxarrow/.style={arrow, gray!55},
    phase/.style={font=\sffamily\small\scshape, text=gray!45!black},
]

\node[font=\sffamily\normalsize\bfseries, anchor=west] at (-3.6, 4.0) {Two-Model Pipeline};
\node[phase, anchor=west] at (-3.6, 3.4) {ColQwen3.5 + Qwen3.5};

\node[databox] (t-docs) at (-2.8, 2.4) {Indexing};
\node[modelbox, minimum width=2.2cm] (t-ret-idx) at (0.0, 2.4)
    {ColQwen3.5\\{\sffamily\scriptsize(4B params)}};
\node[databox, text width=2.0cm] (t-index) at (2.8, 2.4)
    {Index\\(page embeddings)};

\draw[arrow] (t-docs) -- (t-ret-idx);
\draw[arrow] (t-ret-idx) -- (t-index);

\node[databox] (t-query) at (-2.8, 0.2) {Query};
\node[modelbox, minimum width=2.0cm] (t-ret-q) at (0.0, 0.2)
    {ColQwen3.5\\Retrieval};
\node[databox] (t-topk) at (2.6, 0.2) {Top-$k$\\Pages};
\node[genbox, minimum width=2.2cm] (t-llm) at (5.2, 0.2)
    {Qwen3.5\\{\sffamily\scriptsize(4B params)}};
\node[databox] (t-ans) at (7.8, 0.2) {Answer};

\draw[arrow] (t-query) -- (t-ret-q);
\draw[arrow] (t-ret-q) -- (t-topk);
\draw[arrow] (t-topk) -- (t-llm);
\draw[arrow] (t-llm) -- (t-ans);
\draw[idxarrow] (t-index.south) -- ++(0,-0.55) -| (t-ret-q.north);

\node[draw=gray!40, rounded corners=2pt, fill=none, inner sep=5pt,
      text width=3.6cm, align=center]
    at (10.8, 1.3) {
    {\sffamily\footnotesize\bfseries $\sim$8.1B total params}\\[1pt]
    {\sffamily\scriptsize\color{gray!55!black} 28{,}850~MB peak VRAM}\\
    {\sffamily\scriptsize\color{gray!55!black} 2 models in GPU memory}
};

\draw[gray!30, line width=0.6pt] (-3.8, -1.2) -- (12.8, -1.2);

\node[font=\sffamily\normalsize\bfseries, anchor=west] at (-3.6, -1.7) {Single-Model Pipeline};
\node[phase, anchor=west] at (-3.6, -2.3) {\ours{}};

\node[databox] (h-docs) at (-2.8, -3.3) {Indexing};

\node[hydrabox, minimum width=3.4cm] (h-hydra-idx) at (0.0, -3.3) {};
\node[font=\sffamily\small\bfseries] at (0.0, -2.95) {\ours{} {\sffamily\scriptsize(4B params)}};
\node[retlabel] at (0.0, -3.55) {Retrieval head};

\node[databox, text width=2.0cm] (h-index) at (3.2, -3.3)
    {Index\\(page embeddings)};

\draw[arrow] (h-docs) -- (h-hydra-idx);
\draw[arrow] (h-hydra-idx) -- (h-index);

\node[databox] (h-query) at (-2.8, -5.5) {Query};

\node[hydrabox, minimum width=2.6cm] (h-hydra-ret) at (0.0, -5.5) {};
\node[font=\sffamily\small\bfseries] at (0.0, -5.15) {\ours{}};
\node[retlabel] at (0.0, -5.7) {Retrieval head};

\node[databox] (h-topk) at (2.6, -5.5) {Top-$k$\\Pages};

\node[hydrabox, minimum width=2.6cm] (h-hydra-gen) at (5.2, -5.5) {};
\node[font=\sffamily\small\bfseries] at (5.2, -5.15) {\ours{}};
\node[genlabel] at (5.2, -5.7) {Generation head};

\node[databox] (h-ans) at (7.8, -5.5) {Answer};

\draw[arrow] (h-query) -- (h-hydra-ret);
\draw[arrow] (h-hydra-ret) -- (h-topk);
\draw[arrow] (h-topk) -- (h-hydra-gen);
\draw[arrow] (h-hydra-gen) -- (h-ans);
\draw[idxarrow] (h-index.south) -- ++(0,-0.55) -| (h-hydra-ret.north);

\node[draw=gray!40, rounded corners=2pt, fill=none, inner sep=5pt,
      text width=3.6cm, align=center]
    at (10.8, -4.0) {
    {\sffamily\footnotesize\bfseries 4B total params}\\[1pt]
    {\sffamily\scriptsize\color{gray!55!black} 10{,}770~MB peak VRAM}\\
    {\sffamily\scriptsize\color{gray!55!black} 1 model in GPU memory}
};

\end{tikzpicture}%
}
\caption{%
RAG pipeline comparison.
\textbf{Top:} A conventional two-model pipeline (ColQwen3.5 for retrieval + Qwen3.5 for generation) holds both models co-resident in GPU memory ($\sim$8.1B parameters, 28{,}850~MB peak VRAM).
\textbf{Bottom:} \ours{} uses a single 4B-parameter model for both indexing (retrieval head for embeddings) and querying (retrieval head finds top-$k$ pages, generation head answers from them). Both heads share one model in GPU memory, reducing peak VRAM to 10{,}770~MB (savings quantified in \Cref{fig:memory-savings}).
Solid blue borders = retrieval; red borders = generation.
}
\label{fig:rag-pipeline}
\end{figure*}

\section{Related Work}
\label{sec:related}
\paragraph{Unified embedding and generation.}
GritLM~\citep{muennighoff2025gritlm} showed that a single LLM can perform both embedding and generation by alternating between objectives during full fine-tuning, switching between bidirectional and causal attention masks at inference.
\mbox{OneGen~\citep{zhang2024onegen}} unified both in a single forward pass by allocating special retrieval tokens whose hidden states serve as query embeddings during autoregressive generation.
Both remain text-only and use dense single-vector embeddings rather than multi-vector late interaction.

\paragraph{Unified retrieval and generation for visual documents.}
SV-RAG~\citep{chen2025svrag} trains two separate LoRA adapters on a shared frozen MLLM backbone: one converts the model into a ColBERT-style multi-vector retriever, the second fine-tunes it for QA generation, with adapters swapped at inference.
URaG~\citep{shi2026urag} inserts a lightweight retrieval module at an intermediate transformer layer, exploiting the observation that early layers distribute attention broadly while deeper layers concentrate on evidence pages; irrelevant pages are pruned mid-forward-pass, achieving retrieval and generation in a single pass.
VDocRAG~\citep{tanaka2025vdocrag} pre-trains a VLM with both retrieval and generation objectives but deploys separate components at inference.
VisRAG~\citep{yu2024visrag} uses VLMs for both tasks as a two-stage pipeline with separately fine-tuned models.

\ours{} differs from SV-RAG in requiring \emph{one} adapter and \emph{no generation training}: disabling the retrieval adapter exactly recovers the base model's generation capability.
It differs from URaG in producing a standalone ColBERT retriever that can be deployed independently of the generation pathway, rather than coupling retrieval to an intermediate layer of the generation forward pass.

\paragraph{LoRA as an inference-time switch.}
ColQwen2\_4RAG~\citep{oprea2025colqwen24rag} showed that toggling ColQwen2's LoRA adapters on and off switches the same Qwen2-VL backbone between retrieval and generation modes, demonstrating the core mechanism in an application context. Our contribution relative to that work is a systematic characterisation of failure modes that \emph{can} occur under different training setups (\Cref{sec:requirements}), a controlled-baseline retrieval evaluation, and a comparison against joint training.
More broadly, aLoRA~\citep{alora2025} invokes different LoRA adapters at different RAG pipeline stages with KV-cache reuse, MeteoRA~\citep{meteora2025} embeds multiple task-specific LoRA adapters with per-token gating, and S-LoRA~\citep{sheng2024slora} provides serving infrastructure for concurrent adapter selection.
Hydra differs from these approaches in requiring no generation training and providing the systematic failure-mode characterisation in \Cref{sec:requirements}.

\paragraph{Scope of comparison.}
We build on ColQwen3.5~\citep{colqwen35},\footnote{ColQwen3.5 is a public model (\texttt{athrael-soju/colqwen3.5-4.5B-v3}) by the present author; we cite it as the prior single-head retrieval reference on the Qwen3.5 family on which \ours{} is constructed. The \ours{}-4B retrieval comparison in \Cref{sec:experiments} is against a same-recipe matched-seed re-train rather than against the public ColQwen3.5-v3 release, to isolate the dual-head mechanism from training-data and step-count differences.} which adapts Qwen3.5~\citep{qwen35} for ColBERT-style late-interaction retrieval over patch embeddings~\citep{khattab2020colbert}.
Our evaluation is scoped to this family of vision-first, multi-vector models; single-vector and hybrid text-vision approaches use a different retrieval mechanism. The controlled-baseline comparison in \Cref{sec:experiments} is the load-bearing retrieval evidence.

\section{Method}
\label{sec:method}
\subsection{Architecture Overview}

\ours{} consists of a single ColQwen3.5 model (Qwen3.5~\citep{qwen35} augmented with a linear projection head \texttt{custom\_text\_proj}: $\mathbb{R}^{d} \rightarrow \mathbb{R}^{320}$), plus two output pathways (\Cref{fig:architecture}):

\begin{enumerate}[nosep]
    \item \textbf{Retrieval head}: The \texttt{custom\_text\_proj} projection, producing $L_2$-normalized 320-dim multi-vector embeddings for ColBERT-style late-interaction scoring.
    \item \textbf{Generation head}: The base model's \texttt{lm\_head} ($\mathbb{R}^{d} \rightarrow \mathbb{R}^{|V|}$), producing logits over the vocabulary for autoregressive decoding.
\end{enumerate}

A single LoRA adapter ($r{=}32$, $\alpha{=}32$, dropout 0.05) is applied to all language model projection layers (\texttt{q\_proj}, \texttt{k\_proj}, \texttt{v\_proj}, \texttt{o\_proj}, \texttt{gate\_proj}, \texttt{up\_proj}, \texttt{down\_proj}, plus the linear-attention \texttt{in\_proj\_*} and \texttt{out\_proj}) and the \texttt{custom\_text\_proj}, \emph{excluding} the vision encoder. The full target-module list is in \Cref{sec:training}.
The vision encoder remains frozen, ensuring identical visual features in both modes.

\subsection{Mode Switching}

The two heads are activated by toggling two controls:

\paragraph{Retrieval mode (embedding).}
The LoRA adapter is enabled, and full-attention layers are patched to bidirectional attention.
Specifically, for each full-attention layer, we replace the causal attention mask $\mathbf{M}_{\text{causal}}$ with a bidirectional mask $\mathbf{M}_{\text{bidir}}$:
\begin{equation}
    \mathbf{M}_{\text{bidir}}[i,j] = \begin{cases}
        0 & \text{if positions } i \text{ and } j \text{ are both valid (non-padding)} \\
        -\infty & \text{otherwise}
    \end{cases}
\end{equation}
This is implemented by extracting the diagonal of the 4D causal mask to identify valid positions, then constructing a symmetric mask where all valid positions attend to each other.
Sliding-window layers are left unchanged, as their local attention pattern is compatible with both modes.
The forward pass produces hidden states that are projected through \texttt{custom\_text\_proj} and $L_2$-normalized to yield multi-vector embeddings.

\paragraph{Generation mode.}
The LoRA adapter is disabled, restoring the base model weights ($W_{\text{adapted}} - BA = W_{\text{base}}$).
Full-attention layers revert to their original causal attention.
The forward pass produces hidden states that are projected through the base \texttt{lm\_head} for greedy autoregressive decoding.

This mode switching happens per call, with no weight copying or model reloading (\Cref{alg:mode-switching}).

\begin{algorithm}[h]
\caption{Mode switching in \ours{} (high-level toggle). The body of the \textsc{Generate} call additionally invokes the base model's \texttt{forward} directly to enable KV-caching; see Requirement~3 in \S\ref{sec:requirements}.}
\label{alg:mode-switching}
\begin{algorithmic}[1]
\STATE \textbf{function} \textsc{Embed}(images)
\STATE \quad Enable LoRA adapter layers
\STATE \quad Set full-attention layers to bidirectional
\STATE \quad \textbf{return} \texttt{custom\_text\_proj}(forward(images))
\STATE
\STATE \textbf{function} \textsc{Generate}(image, prompt)
\STATE \quad Disable LoRA adapter layers
\STATE \quad Restore causal attention on full-attention layers
\STATE \quad \textbf{return} autoregressive\_decode(\texttt{lm\_head}, forward(image, prompt))
\end{algorithmic}
\end{algorithm}

\subsection{Design Rationale: Retrieval-Only Training}

Prior approaches to unified retrieval and generation (GritLM~\citep{muennighoff2025gritlm} via joint training, SV-RAG~\citep{chen2025svrag} via dual adapters) assume that generation capability must be explicitly trained or preserved.
We show this is unnecessary when using LoRA.

Let $W_{\text{base}}$ denote the frozen base model weights, $B, A$ the LoRA matrices (whose product $BA$ gives the low-rank update), and $\phi_{\text{proj}}$ the \texttt{custom\_text\_proj} parameters.
Retrieval training optimizes $BA$ and $\phi_{\text{proj}}$ via contrastive loss while $W_{\text{base}}$ (including \texttt{lm\_head}) remains frozen.
At generation time, we disable LoRA and use $W_{\text{base}}$ directly.
Since $W_{\text{base}}$ was never modified, the generation capability is \emph{equivalent} to the pretrained VLM at the weight level (see \Cref{sec:generation} for empirical verification).

The ablation in \Cref{sec:ablation} confirms this systematically: joint training provides no measurable benefit.
The LoRA toggling approach is simpler: the base model's weights are recovered \emph{exactly}, yielding generation with no degradation under greedy decoding (\Cref{sec:generation}).

\subsection{Operating Constraints for Dual-Head Generation}
\label{sec:requirements}

LoRA's additive structure makes generation equivalence achievable in principle: in any layer where LoRA is the only adapter, disabling it returns the layer to its pre-LoRA state.
In practice, two mechanisms can corrupt the generation pathway during retrieval fine-tuning (Requirements~1--2 below), and a third constraint determines whether generation is fast enough to be practical (Requirement~3).
Our contribution is the identification of these failure modes; \ours{} is structured so that Requirements~1--2 do not arise, and Requirement~3 is addressed in the decode loop.

\paragraph{Requirement 1: Attention mode restoration (inference-time correctness).}
Retrieval training patches full-attention layers to bidirectional attention.
This is an inference-time setting rather than a stored weight: if the patch is not reverted before generation, autoregressive decoding fails because the model can attend to future tokens during prefill, breaking the causal structure that left-to-right generation depends on.
In Qwen3.5's hybrid architecture, only ``full\_attention'' layers (as opposed to sliding-window layers) require this patching, since sliding-window layers use a fixed local window that is compatible with both modes.
Our implementation stores both the original causal and patched bidirectional forward functions per layer, switching between them at mode-toggle time.

\paragraph{Requirement 2: Base model \texttt{lm\_head} preservation.}
The \texttt{lm\_head} used for generation must be the \emph{original} base model's \texttt{lm\_head}, loaded separately from the pretrained checkpoint.
Although LoRA leaves $W_{\text{base}}$ frozen in principle, in practice the \texttt{lm\_head} can be corrupted during training through two mechanisms we identified empirically.
First, when \texttt{lm\_head} shares tied weights with the input embedding layer~\citep{press2017tying}, gradients from the embedding propagate to \texttt{lm\_head} even though it is not a LoRA target.
Second, failing to set \texttt{requires\_grad=False} on \texttt{lm\_head} allows PyTorch DDP to accumulate and synchronize gradients for it even when no optimizer group updates it, causing bf16 numerical drift over thousands of steps.
Our specific setup sidesteps both failure modes structurally.
First, ColQwen3.5 inherits from \texttt{Qwen3\_5Model} rather than \texttt{Qwen3\_5ForConditionalGeneration}, so the retrieval-trained checkpoint carries no \texttt{lm\_head} tensor at all and the bitwise audit's ``426/426 stored tensors'' set excludes \texttt{lm\_head} by construction.
Second, the token embedding table \texttt{embed\_tokens} is neither matched by the LoRA adapter's \texttt{target\_modules} nor listed in \texttt{modules\_to\_save}, so it remains byte-identical to the vanilla base through training.
Since Qwen3.5 declares \texttt{lm\_head.weight} as tied to \texttt{embed\_tokens.weight}~\citep{press2017tying} via its \texttt{\_tied\_weights\_keys} mapping, the weight matrix needed for generation is already resident inside the retrieval model.
A thin \texttt{nn.Linear} instantiated on PyTorch's meta device and then aliased to \texttt{embed\_tokens.weight} provides the generation head at no additional memory cost. Outputs are byte-identical to loading a separate base-model instance for \texttt{lm\_head} extraction, verified on the 50-sample audit in \Cref{sec:kv-isolation}.

\paragraph{Requirement 3: KV-cache-aware generation (efficiency).}
Without KV-cache, each token generation step requires a full forward pass including vision encoder processing of pixel values.
We implement KV-cache-aware generation: pixel values are processed on the first forward step, and subsequent steps reuse cached key-value pairs.
On a single B200, forcing a 128-token decode on eight DocVQA samples, the KV-cache-aware path averages 3.06\,s per sample (23.9\,ms/tok) versus 23.38\,s per sample (182.7\,ms/tok) for the re-forward-every-step path, an order-of-magnitude speedup (mean $7.65\times$, median $5.78\times$ on this small sample; \texttt{results/efficiency/kv\_cache\_bench/report.json}).
The mean/median gap reflects per-sample variance over $n{=}8$; the headline take-away is that re-running the vision encoder per token is a constant-factor regression on the order of $\sim$6$\times$ rather than a precise multiplier.
This requires calling the base model's forward pass directly (bypassing ColQwen3.5's wrapper, which does not support \texttt{use\_cache=True}) and manually managing the attention mask extension at each step.

\section{Training}
\label{sec:training}
Only the retrieval head is trained.
We use standard ColPali-engine training~\citep{faysse2025colpali} with the ColBERT contrastive loss.

\subsection{Training Data}

We combine four visual document retrieval datasets:
\begin{itemize}[nosep]
    \item \texttt{vidore/colpali\_train\_set}~\citep{faysse2025colpali}
    \item \texttt{openbmb/VisRAG-Ret-Train-Synthetic-data}~\citep{yu2024visrag}
    \item \texttt{openbmb/VisRAG-Ret-Train-In-domain-data}~\citep{yu2024visrag}
    \item \texttt{llamaindex/vdr-multilingual-train}~\citep{vdrmultilingual2024} (en, de, es, fr, it)
\end{itemize}

Each sample consists of a text query paired with a positive document page image; all datasets are publicly available for research use.
After filtering null-query rows in the multilingual set, the concatenated training mix contains 760{,}984 pairs.
\ours{}-0.8B is trained on the \texttt{vidore/colpali\_train\_set} split only (118{,}195 pairs) as a smaller-scale instantiation of the same recipe.
Evaluation uses the test split of \texttt{vidore/colpali\_train\_set} plus the ViDoRe V1/V2/V3 MTEB suites.

\subsection{Training Configuration}

Both \ours{}-4B and \ours{}-0.8B share the following recipe:
\begin{itemize}[nosep]
    \item \textbf{Loss}: ColBERT loss~\citep{khattab2020colbert}, temperature $\tau = 0.02$, in-batch negatives.
    \item \textbf{LoRA}: $r{=}32$, $\alpha{=}32$, dropout $= 0.05$. Applied to all LM projection layers (\texttt{q\_proj}, \texttt{k\_proj}, \texttt{v\_proj}, \texttt{o\_proj}, \texttt{gate\_proj}, \texttt{up\_proj}, \texttt{down\_proj}, plus the linear-attention \texttt{in\_proj\_*} and \texttt{out\_proj}) and the \texttt{custom\_text\_proj}; vision encoder frozen. \texttt{custom\_text\_proj} is additionally listed in \texttt{modules\_to\_save} to record the full base weight of the projection head rather than only the LoRA delta.
    \item \textbf{Retrieval projection}: $\mathbb{R}^{d}\to\mathbb{R}^{320}$ for \ours{}-4B ($d{=}2560$) and $\mathbb{R}^{d}\to\mathbb{R}^{128}$ (engine default) for \ours{}-0.8B.
    \item \textbf{Bidirectional attention}: 8 full-attention layers (out of 32 total transformer layers) patched during training; the remaining 24 sliding-window layers are left unchanged.
    \item \textbf{Optimizer}: AdamW, learning rate $5\times 10^{-5}$, cosine schedule with 2.5\% warmup, weight decay $0.0$.
    \item \textbf{Batch}: \ours{}-4B uses per-device batch 36 across 7 GPUs (effective 252) for 3{,}020 steps (1 epoch). \ours{}-0.8B uses per-device batch 32 across 7 GPUs (effective 224) for 528 steps (1 epoch). bf16 throughout; gradient checkpointing on.
    \item \textbf{Seed}: 42 for both public releases.
\end{itemize}

All results reported are from single training runs per model; the baseline comparison at the same recipe (seed 123) is in \Cref{sec:experiments}.

\section{Experiments}
\label{sec:experiments}
\subsection{Retrieval: ViDoRe Benchmarks}

We evaluate retrieval performance on three ViDoRe benchmark suites: V1~\citep{faysse2025colpali} (10 tasks spanning arxiv papers, forms, tables, and synthetic documents), V2~\citep{mace2025vidore2} (4 tasks: biomedical, ESG, and economics reports), and V3~\citep{loison2026vidore3} (8 multilingual tasks across computer science, energy, finance, HR, industrial, pharmaceuticals, and physics domains), for 22 tasks in total.
Evaluation uses the Massive Text Embedding Benchmark (MTEB) framework~\citep{muennighoff2023mteb} (v2.10.13) with MaxSim scoring.
\Cref{tab:vidore} reports average normalized Discounted Cumulative Gain at rank 5 (\ndcg{}) across all three suites; per-task breakdowns are in \Cref{sec:vidore-pertask} (Appendix).

\begin{table}[t]
\centering
\caption{ViDoRe retrieval performance at both scales. (a) \ours{}-4B paired against a same-recipe single-head baseline ($r{=}32$, seed 123, identical data and step count). (b) \ours{}-0.8B. Per-task results in \Cref{sec:vidore-pertask,sec:scaling-pertask} (Appendix).}
\label{tab:vidore-combined}

\begin{subtable}[t]{0.58\linewidth}
\centering
\caption{\ours{}-4B (22 tasks)}
\label{tab:vidore}
\small
\begin{tabular}{@{}lcccc@{}}
\toprule
Suite & Tasks & \ours{}-4B & Baseline & $\Delta$ \\
\midrule
ViDoRe V1 & 10 & 0.9082 & 0.9105 & $-$0.0023 \\
ViDoRe V2 & 4  & 0.5697 & 0.5612 & $+$0.0085 \\
ViDoRe V3 & 8  & 0.5639 & 0.5641 & $-$0.0002 \\
\midrule
All 22 & 22 & \textbf{0.7215} & 0.7210 & $+$0.0005 \\
\bottomrule
\end{tabular}

\smallskip
{\scriptsize Paired $t$-test $p{=}0.89$; 95\% CI on $\Delta$: $[{-}0.006,{+}0.007]$ (TOST equivalence at $\pm$1\,pp).}
\end{subtable}\hfill
\begin{subtable}[t]{0.38\linewidth}
\centering
\caption{\ours{}-0.8B (22 tasks)}
\label{tab:scaling-vidore}
\small
\begin{tabular}{@{}lcc@{}}
\toprule
Suite & Tasks & \ndcg{} \\
\midrule
ViDoRe V1 & 10 & 0.8560 \\
ViDoRe V2 & 4  & 0.5317 \\
ViDoRe V3 & 8  & 0.4434 \\
\midrule
All 22 & 22 & \textbf{0.6470} \\
\bottomrule
\end{tabular}

\smallskip
{\scriptsize~}
\end{subtable}
\end{table}

V1 is saturated (mean 0.9082, all 10 tasks $\geq$0.65); V2 and V3 drop as expected on the harder suites.
We trained a single-head retrieval-only baseline on the same Qwen3.5-4B base with an identical recipe (3{,}020 steps, seed 123) to isolate the effect of the dual-head mechanism from recipe choice.
Across 22 tasks, mean $\Delta{=}{+}0.0005$ with a 95\% CI of $[{-}0.006,{+}0.007]$, excluding effects larger than $\sim$0.7\,pp in either direction (paired $t$-test $p{=}0.89$, Wilcoxon $p{=}1.00$, Cohen's $d{=}0.03$). This satisfies a TOST~\citep{schuirmann1987tost} two-one-sided test against a $\pm$1\,pp margin.
\ours{} wins 9 tasks, loses 11, and ties 2; no per-task delta exceeds 5 percentage points (max $+4.97$\,pp on ESG Reports HL). The dual-head modification therefore adds generation capability without measurable retrieval cost relative to a matched single-head baseline. Equivalent retrieval-vs-baseline measurements at the 0.8B scale are future work; the 0.8B variant is reported in isolation against its own scaled training run (\Cref{tab:scaling-vidore}).

\subsection{Scaling Validation: \ours{}-0.8B}
\label{sec:scaling}

To test whether the dual-head mechanism generalises across scale, we train a smaller instantiation on the same Qwen3.5 family: \ours{}-0.8B (dim 128, trained on \texttt{vidore/colpali\_train\_set} only, 528 steps, same LoRA recipe).
\Cref{tab:scaling-vidore} reports its retrieval performance.
The 0.8B variant drops ${\approx}5$\,pp on V1 and ${\approx}4$--$12$\,pp on the harder suites relative to 4B, a monotonic scaling trend consistent with the base VLM's smaller capacity. All per-task scores are in \Cref{sec:scaling-pertask} (Appendix). The efficiency results below record 62.7\% savings at 4B and 59.1\% at 0.8B; the ratio is stable across scale.

\subsection{Generation Quality}
\label{sec:generation}

Since the generation head uses the unmodified base VLM with LoRA disabled, generation quality should be equivalent to the pretrained Qwen3.5.

\paragraph{Bitwise weight recovery.}
As a direct test of the claim that $W_{\text{adapted}}{-}BA{=}W_{\text{base}}$ under LoRA, we audit every language-model weight tensor in the \ours{} stack (LoRA disabled) against a freshly-loaded \texttt{Qwen3\_5ForConditionalGeneration}. Three invariants hold at the 4B scale: (i) the adapter config targets only retrieval-side modules; (ii) \texttt{adapter\_model.safetensors} contains no \texttt{lm\_head} or \texttt{embed\_tokens} tensors (the only non-LoRA entries are the retrieval-side \texttt{custom\_text\_proj} pair); and (iii) \textbf{426 of 426 language-model weight tensors are byte-for-byte identical} to the vanilla base. The 0.8B variant passes the same audit with zero mismatches. Scripts and report JSONs are in \texttt{scripts/test\_gen\_equivalence\_4b.py} (4B) and \texttt{scripts/test\_gen\_equivalence.py} (0.8B), with outputs under \texttt{results/generation\_equivalence/} in each model repo.

\paragraph{Decoded-output equivalence.}
We use greedy decoding ($T{=}0$) to produce deterministic outputs across four VQA benchmarks (DocVQA~\citep{mathew2021docvqa}, ChartQA~\citep{masry2022chartqa}, InfoVQA~\citep{mathew2022infographicvqa}, and TextVQA~\citep{singh2019textvqa}), totalling 15,301 samples, using Average Normalized Levenshtein Similarity (ANLS)~\citep{biten2019anls}. Per-benchmark ANLS deltas are small and sign-mixed: DocVQA $+$0.0041, ChartQA $+$0.0000, InfoVQA $-$0.0043, TextVQA $+$0.0008 (max $|\Delta|{=}0.0043$; \Cref{tab:generation} in Appendix). The residual deltas come from differences between \ours{}'s custom KV-cache generation loop and \texttt{transformers}' \texttt{.generate()} path (final-norm application, EOS token set, logit post-processing); both paths apply identical LM weights, as verified bitwise above. The KV-cache isolation test (\Cref{sec:kv-isolation}) is a separate self-consistency check across mode switches (50/50 byte-identical outputs, zero embedding drift).

\subsection{Ablation: Joint Training vs.\ LoRA Toggle}
\label{sec:ablation}

An alternative to \ours{}'s retrieval-only training is GritLM-style joint training~\citep{muennighoff2025gritlm}, which alternates between embedding and generation batches during fine-tuning.
For this ablation we train a joint model on the same Qwen3.5-4B base using alternating batches (80\% ColBERT loss, 20\% cross-entropy on LLaVA-Instruct VQA data~\citep{liu2023llava}), with LoRA $r{=}16$, $\alpha{=}64$, dropout $0.197$. Batch size is 32 for the joint model vs.\ 112 for the matched retrieval-only partner; interleaving generation batches inflates memory.
We evaluate both models in three inference modes: LoRA-on retrieval, LoRA-off generation, and LoRA-on generation (the mode GritLM-style training is designed to enable).

\begin{table}[t]
\centering
\caption{Three-mode comparison of \ours{} (retrieval-only training) vs.\ GritLM-style joint training, both at $r{=}16$. Retrieval: 9 ViDoRe V1 tasks. Generation: DocVQA validation ($n{=}200$). Both models use the same base and LoRA config; batch size differs (32 for the joint model vs.\ 112 for the \ours{} partner) due to the additional memory cost of interleaving generation batches. The LoRA-off causal row reports a single shared measurement: with the LoRA adapter disabled the two checkpoints share the same base \texttt{lm\_head} and \texttt{embed\_tokens}, so the LoRA-off forward pass is identical for both. Per-checkpoint LoRA-off ANLS for the \ours{} partner is therefore not separately measured; the column-spanning entry reflects this. Despite different adapter weights when LoRA is on (max element-wise diff: 0.50), the two functional modes are equivalent on retrieval and on LoRA-off generation; the mode that joint training was designed to unlock (LoRA-on causal) fails.}
\label{tab:ablation}
\small
\begin{tabular}{@{}lcc@{}}
\toprule
Inference Mode & \ours{} & GritLM-style \\
\midrule
LoRA-on, bidirectional (retrieval)   & 0.8842 \ndcg{}           & 0.8893 \ndcg{}           \\
LoRA-off, causal (shared base generation)$^\ddagger$  & \multicolumn{2}{c}{0.543 / 0.561 ANLS, 76.5\% match} \\
LoRA-on, causal (target of joint training)  & N/A                      & \textit{image-blind}$^\dagger$ \\
\bottomrule
\end{tabular}

\smallskip
\textsuperscript{$\dagger$}See text. The $n{=}200$ subset is sufficient to detect this catastrophic failure mode; the test is binary (does the model condition on image content at all?).
\textsuperscript{$\ddagger$}Cell measured against the GritLM-style adapter-off path; the \ours{} partner's adapter-off measurement is not run independently because, with the same base weights and LoRA disabled, the two paths are mathematically identical.
\end{table}

\Cref{tab:ablation} summarizes the results.
On the LoRA-on retrieval mode and LoRA-off generation mode, the two training approaches are equivalent within experimental noise, despite large adapter-weight differences (max element-wise diff 0.50).

The notable finding is that LoRA-on generation, the mode that joint training was designed to enable, fails entirely.
On DocVQA ($n{=}200$, $T{=}0$), the jointly-trained model produces a single token (``The'') with probability $p{=}0.91$ regardless of image content, unable to condition on visual input.
LoRA toggling is therefore structurally necessary at this rank: the low-rank subspace cannot simultaneously serve both attention modes.
GritLM's full fine-tuning successfully supports both modes~\citep{muennighoff2025gritlm}, so the LoRA-on causal failure is a low-rank constraint rather than a fundamental property of bidirectional attention. In-batch negatives dominate ColBERT-loss signal, so the $\sim$3.5$\times$ batch difference is a confound on the retrieval delta (0.8893 vs.\ 0.8842 nDCG@5, GritLM slightly ahead despite the smaller batch). The retrieval equivalence claim rests on this being within experimental noise; the single-token collapse does not.

\subsection{Efficiency}

We measure the practical overhead of the single-model architecture across two scales. All measurements are from the audit scripts released alongside the public model weights; report JSONs are in the respective \texttt{results/mode\_switch\_vram/} directories.

\paragraph{Memory.}
A two-model deployment of ColQwen3.5 + Qwen3.5 (both co-resident in GPU memory, as a serving deployment that wishes to amortise model load and parallelise embed/generate would) peaks at 28.85\,GB at the 4B scale, while \ours{}-4B peaks at 10.77\,GB over the same cycle, a 62.7\% reduction (\Cref{fig:memory-savings}).
The 0.8B variant shows 59.1\% savings, confirming the ratio is stable across scale.
Hydra generation (4B: 10.77\,GB, 0.8B: 2.37\,GB) is the peak-memory mode in both cases since both weight branches are resident.\footnote{At 0.8B, Hydra retrieval (1.90\,GB) fits below vanilla base generation (2.02\,GB) because the ColQwen3.5 shell has no \texttt{lm\_head} tensor until it is aliased on demand; at 4B the effect reverses (retrieval 9.57\,GB vs.\ vanilla base 9.46\,GB) because the larger \texttt{custom\_text\_proj} (dim 320) outweighs the absent \texttt{lm\_head} alias. The 28.85\,GB baseline is the simultaneous-residency serving configuration that \ours{}'s single-model design most directly displaces; a lazy-loading pipeline would trade reload latency for lower steady-state peak.}

\begin{figure}[t]
\centering
\definecolor{steelblue}{RGB}{55,126,184}
\definecolor{warmred}{RGB}{200,80,80}
\begin{tikzpicture}[
    ticklabel/.style={font=\sffamily\scriptsize, text=black!70},
    axislabel/.style={font=\sffamily\footnotesize, text=black!80},
    barlabel/.style={font=\sffamily\scriptsize\bfseries, anchor=south},
    grouplabel/.style={font=\sffamily\footnotesize\bfseries, text=black!85, anchor=north},
    delta/.style={font=\sffamily\small\bfseries, text=steelblue!30!black},
]

    \def\yscale{0.11}          
    \def\chartw{11.5}          
    \def\barw{0.7}             
    \def\bargap{0.1}           
    \def\grpA{3.2}             
    \def\grpB{8.3}             
    \def\ymax{32}              

    \draw[gray!55] (0, 0) -- (\chartw, 0);
    \draw[gray!55] (0, 0) -- (0, \ymax*\yscale + 0.1);
    \foreach \y in {0,8,16,24,32} {
        \draw[gray!25, dashed] (0.02, \y*\yscale) -- (\chartw, \y*\yscale);
        \draw[gray!55] (-0.08, \y*\yscale) -- (0, \y*\yscale);
        \node[ticklabel, anchor=east] at (-0.12, \y*\yscale) {\y};
    }
    \node[axislabel, rotate=90, anchor=south] at (-0.85, \ymax*\yscale / 2)
        {Peak VRAM (GB)};

    \pgfmathsetmacro{\ared}{5.79*\yscale}
    \pgfmathsetmacro{\ablue}{2.37*\yscale}
    \pgfmathsetmacro{\Aredx}{\grpA - \bargap/2 - \barw}
    \pgfmathsetmacro{\Abluex}{\grpA + \bargap/2}

    \fill[warmred!12] (\Aredx, 0) rectangle (\Aredx+\barw, \ared);
    \draw[warmred, line width=0.9pt] (\Aredx, 0) rectangle (\Aredx+\barw, \ared);
    \node[barlabel, text=warmred!40!black] at (\Aredx+\barw/2, \ared) {5.79};

    \fill[steelblue!12] (\Abluex, 0) rectangle (\Abluex+\barw, \ablue);
    \draw[steelblue, line width=0.9pt] (\Abluex, 0) rectangle (\Abluex+\barw, \ablue);
    \node[barlabel, text=steelblue!40!black] at (\Abluex+\barw/2, \ablue) {2.37};

    \node[grouplabel] at (\grpA, -0.15) {\ours{}-0.8B};

    \draw[gray!65, line width=0.9pt, -{Stealth[length=5pt, width=4pt]}]
        (\Aredx+\barw/2, \ared + 0.18)
        .. controls (\Aredx+\barw/2, \ared + 0.75)
               and (\Abluex+\barw/2, \ared + 0.75) ..
        (\Abluex+\barw/2, \ablue + 0.18);
    \node[delta] at (\grpA, \ared + 0.75) {$-59.1\%$};

    \pgfmathsetmacro{\bred}{28.85*\yscale}
    \pgfmathsetmacro{\bblue}{10.77*\yscale}
    \pgfmathsetmacro{\Bredx}{\grpB - \bargap/2 - \barw}
    \pgfmathsetmacro{\Bbluex}{\grpB + \bargap/2}

    \fill[warmred!12] (\Bredx, 0) rectangle (\Bredx+\barw, \bred);
    \draw[warmred, line width=0.9pt] (\Bredx, 0) rectangle (\Bredx+\barw, \bred);
    \node[barlabel, text=warmred!40!black] at (\Bredx+\barw/2, \bred) {28.85};

    \fill[steelblue!12] (\Bbluex, 0) rectangle (\Bbluex+\barw, \bblue);
    \draw[steelblue, line width=0.9pt] (\Bbluex, 0) rectangle (\Bbluex+\barw, \bblue);
    \node[barlabel, text=steelblue!40!black] at (\Bbluex+\barw/2, \bblue) {10.77};

    \node[grouplabel] at (\grpB, -0.15) {\ours{}-4B};

    \draw[gray!65, line width=0.9pt, -{Stealth[length=5pt, width=4pt]}]
        (\Bredx+\barw/2, \bred + 0.18)
        .. controls (\Bredx+\barw/2, \bred + 0.75)
               and (\Bbluex+\barw/2, \bred + 0.75) ..
        (\Bbluex+\barw/2, \bblue + 0.18);
    \node[delta] at (\grpB, \bred + 0.75) {$-62.7\%$};

    \pgfmathsetmacro{\legy}{-1.15}
    \pgfmathsetmacro{\legMid}{\chartw/2}
    \pgfmathsetmacro{\itemAx}{\legMid - 3.7}
    \pgfmathsetmacro{\itemBx}{\legMid + 0.6}

    \fill[warmred!12] (\itemAx, \legy) rectangle (\itemAx+0.32, \legy+0.22);
    \draw[warmred, line width=0.9pt] (\itemAx, \legy) rectangle (\itemAx+0.32, \legy+0.22);
    \node[font=\sffamily\footnotesize, anchor=west]
        at (\itemAx+0.4, \legy+0.11) {Two-model deployment};

    \fill[steelblue!12] (\itemBx, \legy) rectangle (\itemBx+0.32, \legy+0.22);
    \draw[steelblue, line width=0.9pt] (\itemBx, \legy) rectangle (\itemBx+0.32, \legy+0.22);
    \node[font=\sffamily\footnotesize, anchor=west]
        at (\itemBx+0.4, \legy+0.11) {\ours{} (single model)};

\end{tikzpicture}
\caption{%
Peak GPU memory for a retrieve-then-generate workflow at both scales.
A conventional two-model deployment holds a ColQwen3.5 retriever and a Qwen3.5 generator simultaneously (red); \ours{} holds a single set of weights and toggles the LoRA adapter in place (blue).
The single-model design cuts peak VRAM by \textbf{59.1\%} at the 0.8B scale and \textbf{62.7\%} at the 4B scale; the ratio is stable across backbone size.
Numbers from \texttt{bench\_mode\_switch\_vram\_4b.py} (4B) and \texttt{bench\_mode\_switch\_vram.py} (0.8B).
}
\label{fig:memory-savings}
\end{figure}

\paragraph{Mode-switching latency.}
A full round-trip (retrieval $\rightarrow$ generation $\rightarrow$ retrieval) takes 6.4\,ms mean (median 6.3\,ms; min--max 5.9--7.6\,ms) over 50 iterations, representing 1.9\% of a single generation call (331\,ms) and making it negligible relative to inference (\Cref{fig:switching-latency}). The latency is dominated by the LoRA enable/disable plus attention-mode swap and is not rank-sensitive in our measurements.

\begin{figure}[t]
\centering
\definecolor{steelblue}{RGB}{55,126,184}
\definecolor{lightblue}{RGB}{220,235,246}
\begin{tikzpicture}[scale=1]
    \def\xs{0.03625} 
    \def\barheight{0.8}
    \foreach \t/\lbl in {0/0, 50/50, 100/100, 150/150, 200/200, 250/250, 300/300, 331/331} {
        \draw[gray!50] (\t*\xs, -0.05) -- (\t*\xs, \barheight + 0.05);
        \node[font=\scriptsize\sffamily, anchor=north, text=black!70] at (\t*\xs, -0.05) {\lbl};
    }
    \node[font=\footnotesize\sffamily, anchor=north] at (6.0, -0.7) {Wall time (ms)};
    \fill[lightblue] (0,0) rectangle (331*\xs, \barheight);
    \draw[steelblue!40!black, line width=0.5pt] (0,0) rectangle (331*\xs, \barheight);
    \fill[steelblue] (0,0) rectangle (6.4*\xs, \barheight);
    \node[font=\scriptsize\sffamily\bfseries, text=steelblue!20!black] at (170*\xs, \barheight/2) {331\,ms};
    \draw[thick, -{Stealth[length=5pt]}, steelblue!60!black]
        (1.5, \barheight+0.8) -- (6.4*\xs, \barheight+0.05);
    \node[font=\scriptsize\sffamily\bfseries, text=steelblue!30!black, anchor=south west]
        at (0.6, \barheight+0.6) {\textbf{6.4\,ms} (1.9\%)};
    \begin{scope}[shift={(1.95, -1.6)}]
        \fill[steelblue] (0, 0) rectangle (0.4, 0.25);
        \draw[steelblue!70!black] (0, 0) rectangle (0.4, 0.25);
        \node[font=\footnotesize\sffamily, anchor=west] at (0.5, 0.12)
            {mode-switch round-trip};
        \fill[lightblue] (4.4, 0) rectangle (4.8, 0.25);
        \draw[steelblue!40!black] (4.4, 0) rectangle (4.8, 0.25);
        \node[font=\footnotesize\sffamily, anchor=west] at (4.9, 0.12)
            {single generation call};
    \end{scope}
\end{tikzpicture}
\caption{%
Mode-switching latency is negligible.
A full round-trip (retrieval $\rightarrow$ generation $\rightarrow$ retrieval) takes \textbf{6.4\,ms} mean (median 6.3\,ms; min--max 5.9--7.6\,ms) over 50 iterations---\textbf{1.9\%} of a single generation call (331\,ms).
The switch cost is dominated by enabling/disabling the LoRA adapter plus swapping the attention mode on 8 full-attention layers; both are pointer-level operations rather than matmul.
Source: \texttt{results/efficiency/mode\_switch/mode\_switch\_latency\_report.json} in the public \ours{}-4B repo; measured on a single NVIDIA B200.
}
\label{fig:switching-latency}
\end{figure}

\paragraph{KV-cache state isolation.}\label{sec:kv-isolation}
A shared model raises the concern that internal state from one mode could leak into the other.
We test this with a contamination protocol on 50 DocVQA inputs: compare \emph{single-pass} \ours{} generation against \emph{round-trip} generation sandwiched between two \ours{} embed calls (embed${\,\rightarrow\,}$generate(decoy)${\,\rightarrow\,}$embed${\,\rightarrow\,}$generate-under-test).
Embeddings are bitwise identical across the two embed calls (max element-wise diff${=}0.0$, cosine similarity${=}1.0$) and final-generation outputs are byte-identical to the single-pass baseline in 100\% of cases (50/50). No KV-cache or attention-mode state persists across mode switches. Report JSON: \texttt{results/kv\_cache\_isolation/} in the model repo.

\subsection{Summary}

\begin{table}[t]
\centering
\caption{Summary of \ours{}. Efficiency measured on a single GPU.}
\label{tab:scaling}
\small
\begin{tabular}{@{}lll@{}}
\toprule
Metric & \ours{}-4B & \ours{}-0.8B \\
\midrule
ViDoRe V1 mean \ndcg{} & 0.9082 (10 tasks) & 0.8560 (10 tasks) \\
ViDoRe V2 mean \ndcg{} & 0.5697 & 0.5317 \\
ViDoRe V3 mean \ndcg{} & 0.5639 & 0.4434 \\
Gen equivalence (LM weights bitwise match) & 426/426 & 320/320 \\
Trainable parameters  & 65.75M (1.43\% of 4.60B) & scaled by backbone \\
Peak VRAM (Hydra)     & 10.77\,GB & 2.37\,GB \\
Peak VRAM (two-model) & 28.85\,GB & 5.79\,GB \\
\bottomrule
\end{tabular}
\end{table}

\Cref{tab:scaling} consolidates results across all evaluation dimensions for both public model scales.

\section{Omni-Modal Extension}
\label{sec:omni}
\subsection{Proof of Concept: Omni-Modal Generalization}

To test whether the \ours{} mechanism generalizes beyond a single model family and modality, we apply it, without additional training, to \textbf{Qwen2.5-Omni-3B}~\citep{xu2025qwen25omni}, a multimodal model with native support for image, audio, and video input, as well as text and speech output.
\paragraph{Setup.}
We use \texttt{vidore/colqwen-omni-v0.1}\footnote{\url{https://huggingface.co/vidore/colqwen-omni-v0.1}}, a ColBERT adapter trained on 127K image-text pairs atop the Qwen2.5-Omni-3B backbone using colpali-engine~\citep{faysse2025colpali}.
The adapter was trained on image data only; audio retrieval and video embedding are produced zero-shot through the base model's frozen Whisper audio encoder~\citep{radford2023whisper} and Qwen2.5-Omni's native Thinker--Talker vision stack~\citep{xu2025qwen25omni}.
We apply the \ours{} architecture as-is: LoRA-on with bidirectional attention for retrieval (via \texttt{custom\_text\_proj}, 128-dim embeddings), LoRA-off with causal attention for generation (via the base model's \texttt{lm\_head}).
No additional training is performed.

The model additionally supports speech synthesis via the Qwen2.5-Omni talker module and BigVGAN vocoder~\citep{lee2023bigvgan}, giving \ours{} three inference modes from a single 4.4B-parameter model instance:

\begin{enumerate}[nosep]
    \item \textbf{Retrieval} (LoRA on, bidirectional): ColBERT multi-vector embeddings over images, audio, or video.
    \item \textbf{Text generation} (LoRA off, causal): Autoregressive text conditioned on any input modality.
    \item \textbf{Speech generation} (LoRA off, causal, talker enabled): Spoken answers via thinker--talker--vocoder pipeline.
\end{enumerate}

\paragraph{Image retrieval.}
The model achieves 0.8865 average \ndcg{} on V1 (10 tasks), 0.5353 on V2 (4 tasks), and 0.4907 on V3 (8 tasks). Relative to \ours{}-4B (\Cref{tab:vidore}), the Omni variant lands within $\sim$2\,pp on V1 (0.8865 vs.\ 0.9082), $\sim$3\,pp on V2 (0.5353 vs.\ 0.5697), and lower on V3 ($-$7.3\,pp; 0.4907 vs.\ 0.5639) despite a smaller backbone (3B) and different model family; full per-task results are in \Cref{tab:omni-vidore} (Appendix).

\paragraph{Audio retrieval (zero-shot).}
We evaluate text-to-audio retrieval on AudioCaps~\citep{kim2019audiocaps} ($n{=}500$ test clips, 7--10\,s each at 16\,kHz).
Audio clips are embedded by routing raw waveforms through the Whisper feature extractor~\citep{radford2023whisper} and the shared projection head; captions are embedded as text queries through the same backbone.
Using ColBERT MaxSim scoring over the full $500{\times}500$ similarity matrix, the model achieves R@1\,=\,26.2\%, R@5\,=\,55.6\%, R@10\,=\,69.0\%, and MRR\,=\,40.6\%, with no audio contrastive training, relying entirely on cross-modal transfer through the shared Qwen2.5-Omni backbone.
For reference, supervised audio-text models (e.g., CLAP~\citep{elizalde2023clap}) achieve R@1\,$\approx$\,35--40\% on this benchmark; the gap is expected given zero-shot transfer.

\paragraph{Generation equivalence.}
We evaluate generation preservation on InfographicVQA~\citep{mathew2022infographicvqa} validation ($n{=}2{,}801$) under greedy decoding ($T{=}0$, 128 new tokens) with a short-answer prompt suffix applied identically to both paths to suppress Qwen2.5-Omni's default sentence-form outputs. Base Qwen2.5-Omni-3B and \ours{}-Omni (LoRA off, \texttt{lm\_head} extracted from the thinker) attain identical strict ANLS of \textbf{0.7257} with $\Delta{=}{+}0.0000$ (95\% CI [${+}0.0000$, ${+}0.0000$]) and \textbf{2{,}801/2{,}801 (100.00\%) byte-identical outputs} across the full validation set. The short-answer prompt suffix used here was not used in the \ours{}-4B InfoVQA run (max $|\Delta|{=}0.0043$, 31.06\% exact match; \Cref{tab:generation}), so the two decoding setups differ. The published report is \texttt{results/infovqa\_report.json} in the Omni repo.

\paragraph{Speech generation.}
\ours{}-Omni can also produce spoken answers by routing through the thinker, talker, and BigVGAN vocoder~\citep{lee2023bigvgan} pipeline, producing coherent 24\,kHz speech from the same model instance. Video embeddings are produced by the pipeline as a forward-pass output but not evaluated as retrieval.

\section{Discussion}
\label{sec:discussion}
\paragraph{LoRA as a mode switch.}
The ablation in \Cref{sec:ablation} confirms that LoRA toggling, rather than joint training, is the operative mechanism: GritLM-style training achieves equivalent results but still requires toggling, confirming the additional complexity provides no benefit.

\paragraph{Comparison with prior unified architectures.}
\Cref{tab:arch-comparison} compares \ours{} against prior unified retrieval-generation architectures across key design dimensions.
\ours{} is the only approach that requires no generation training and uses a single adapter; the base model's generation capability is recovered by disabling the adapter rather than being explicitly trained or preserved.

\begin{table}[t]
\centering
\caption{Structural comparison of unified retrieval-generation architectures. \ours{} is the only approach requiring no generation training and a single adapter. VRAM figures for \ours{} are measured at the 4B scale; rows for other systems are structural rather than measured (we did not run identical co-resident benchmarks against SV-RAG, URaG, or GritLM).}
\label{tab:arch-comparison}
\small
\setlength{\tabcolsep}{5pt}
\resizebox{\textwidth}{!}{%
\begin{tabular}{@{}lcccc@{}}
\toprule
Property & \ours{} & SV-RAG & URaG & GritLM \\
\midrule
Adapters needed       & 1                 & 2 (swapped)    & 0 (custom)        & 0 (full FT) \\
Generation training   & None              & Yes            & Yes               & Yes \\
Retriever independence & Yes              & Yes            & No                & N/A \\
Multi-vector retrieval & Yes              & Yes            & Yes               & No \\
Resident backbones    & 1 (10.77\,GB$^*$) & 1 (2 adapters) & 1 (custom module) & 1 \\
\bottomrule
\end{tabular}%
}

\smallskip
{\scriptsize $^*$\,Measured for \ours{} at 4B; rows for other systems are structural rather than measured. SV-RAG~\citep{chen2025svrag} swaps two LoRA adapters on a shared frozen MLLM backbone (\S\ref{sec:related}); only one backbone is resident at a time, structurally comparable to \ours{}. The distinction in this row is the \emph{number of adapters} and \emph{whether generation training is required}, not the number of resident backbones.}
\end{table}

\paragraph{Production deployment considerations.}
\ours{}'s single-model design reduces memory but introduces deployment trade-offs.
LoRA adapters incur measurable throughput overhead in current serving frameworks~\citep{sheng2024slora}.
Additionally, the model cannot serve retrieval and generation requests simultaneously; mode switches serialize these operations at the model level, unlike a two-model deployment that can parallelize them across concurrent queries.
LoRA serving infrastructure (S-LoRA~\citep{sheng2024slora}, vLLM adapter routing) is actively improving, but deployments should evaluate throughput requirements alongside memory constraints.

\paragraph{Limitations.}
\begin{itemize}[nosep]
    \item \textbf{VLM families}: Tested on Qwen3.5 (0.8B, 4B) and Qwen2.5-Omni (3B). While the omni-modal extension (\Cref{sec:omni}) demonstrates generality across model families and modalities, testing on non-Qwen architectures (InternVL, LLaVA) remains future work.
    \item \textbf{Single training run}: Canonical 4B and 0.8B results are each from one training run; variance across seeds is not estimated. The 4B same-recipe single-head baseline (seed 123) used throughout \Cref{sec:experiments} is reported in full per-suite (\Cref{tab:vidore}, \Cref{tab:vidore-pertask}); we do not have a matched-seed baseline for 0.8B.
    \item \textbf{Generation evaluation}: Bitwise equivalence verified for both scales (\Cref{sec:generation}); decoded-output ANLS measurements are reported per-benchmark in the appendix and via the audit scripts released with the model.
    \item \textbf{Audio/video retrieval}: The omni-modal results (\Cref{sec:omni}) are zero-shot; explicit audio and video contrastive training would likely improve performance but is not explored.
    \item \textbf{Ablation scope}: The GritLM-style comparison (\Cref{sec:ablation}) is run at $r{=}16$, $\alpha{=}64$, single seed, 20/80 generation/retrieval mix. The single-token collapse under LoRA-on causal is categorical; the retrieval-equivalence conclusion is bounded by the ablation's scope.
    \item \textbf{Video retrieval}: The omni-modal extension (\Cref{sec:omni}) verifies that the pipeline produces video embeddings but does not evaluate them on retrieval benchmarks. ``Video embedding'' should not be interpreted as ``video retrieval.''
    \item \textbf{End-to-end RAG}: Retrieval and generation are evaluated independently. We do not evaluate the full retrieve-then-generate pipeline (\Cref{fig:rag-pipeline}) end-to-end; combined pipeline quality (e.g., answer accuracy given retrieved context) remains untested.
\end{itemize}

\paragraph{Reproducibility.}
External reproducers should be aware of one packaging-level pitfall in the dependency stack: in the \texttt{colpali-engine} 0.3.15 release we tested against, loading \texttt{ColQwen3\_5} directly from the public Qwen3.5-4B checkpoint random-initialises the language-model weights because the Hub checkpoint's \texttt{model.layers.*} key prefix is not stripped by the package's checkpoint-conversion mapping. The working path is to transplant from \texttt{Qwen3\_5ForConditionalGeneration} before attaching the adapter (see \texttt{test\_gen\_equivalence\_\{4b,08b\}.py} in the released model repos). All published \ours{} checkpoints include the post-transplant state, so \texttt{from\_pretrained} on the released revisions does not require manual patching.

\paragraph{Future work.}
Several directions are promising:
(1)~testing on non-Qwen VLM families (InternVL~\citep{chen2024internvl}, LLaVA~\citep{liu2023llava});
(2)~multi-page cross-attention for document-level reasoning;
(3)~explicit audio and video contrastive training to improve zero-shot retrieval performance;
(4)~adapter composition for additional tasks beyond retrieval and generation;
(5)~collapsing multi-component visual-document RAG stacks into one local VLM, trading specialist-component quality and hosted-API access for deployment locality at a capability cost we do not characterise here.

\paragraph{Broader impact.}
\ours{} can process sensitive documents (medical records, legal filings, financial reports), and the single-model design concentrates both retrieval and generation behind one access point.
This simplifies access control relative to multi-model pipelines, but a compromised model exposes both capabilities simultaneously.
Deployments should enforce document-level permissions and audit query logs accordingly.

\section{Conclusion}
\label{sec:conclusion}
\ours{} demonstrates that a single retrieval-trained LoRA adapter suffices to provide both ColBERT-style document retrieval and autoregressive generation from one VLM instance, with no generation training.
The key practical insight is the characterisation of failure modes that \emph{can} silently corrupt the base model's generation capability under different training setups (weight-tying gradients, DDP synchronization artifacts), together with the observation that the same weight-tying property provides a free recovery path when the architecture leaves \texttt{embed\_tokens} frozen (\Cref{sec:requirements}).

The ablation indicates that joint training does not make the adapted weights support both attention modes simultaneously, so toggling is required regardless. The omni-modal extension shows the mechanism preserves image retrieval and generation equivalence on a non-Qwen3.5 backbone (Qwen2.5-Omni-3B), with audio retrieval emerging zero-shot through the base model's frozen Whisper encoder. The 0.8B sister model confirms the VRAM-savings ratio is stable across scale.

More broadly, LoRA adapters are not merely a training convenience; they are inference-time mode switches.
One model, two heads.

\paragraph{Code and models.}
The full set of artefacts is indexed in a single HuggingFace collection: \url{https://huggingface.co/collections/athrael-soju/hydra-dual-head-retrieval-and-generation}.

Individual model weights are released at \url{https://huggingface.co/athrael-soju/HydraQwen3.5-4B} (Hydra-4B), \url{https://huggingface.co/athrael-soju/HydraQwen3.5-0.8B} (small-scale instantiation, with a live Gradio Space at \url{https://huggingface.co/spaces/athrael-soju/HydraQwen3.5-0.8B-demo}), and \url{https://huggingface.co/athrael-soju/HydraQwen2.5-Omni-3B} (omni extension). The GritLM-style joint-training ablation is at \url{https://huggingface.co/athrael-soju/DualHead-GritLM-Qwen3.5-4B}, and the matched single-head retrieval-only baseline lives inside the Hydra-4B repository under \texttt{baseline/}. Training and evaluation scripts ship inside each model repository under \texttt{scripts/}; per-benchmark results live under \texttt{results/}.

\appendix

\section{Per-Task Retrieval Results: \ours{}-4B}
\label{sec:vidore-pertask}

\begin{table}[H]
\centering
\caption{Per-task retrieval performance of \ours{}-4B and the matched single-head baseline (same recipe, $r{=}32$, seed 123) on ViDoRe V1, V2, and V3 (MTEB v2.10.13, MaxSim scoring).}
\label{tab:vidore-pertask}
\small
\begin{tabular}{@{}lcc@{\hskip 1.5em}lcc@{}}
\toprule
\multicolumn{3}{c}{\textit{ViDoRe V1 (10 tasks)}} & \multicolumn{3}{c}{\textit{ViDoRe V2 \& V3}} \\
\cmidrule(r){1-3} \cmidrule(l){4-6}
Task & \ours{}-4B & Baseline & Task & \ours{}-4B & Baseline \\
\midrule
ArxivQA           & 0.9193 & 0.9223 & \multicolumn{3}{@{}l}{\textit{V2}} \\
DocVQA            & 0.6593 & 0.6570 & BioMedical Lectures  & 0.5939 & 0.6096 \\
InfoVQA           & 0.9328 & 0.9405 & ESG Reports (HL)    & 0.7328 & 0.6831 \\
ShiftProject      & 0.9133 & 0.9233 & ESG Reports         & 0.5207 & 0.5264 \\
SynthDocQA-AI     & 1.0000 & 1.0000 & Economics Reports   & 0.4314 & 0.4256 \\
SynthDocQA-Energy & 0.9776 & 0.9826 & \textbf{V2 Average} & \textbf{0.5697} & \textbf{0.5612} \\
SynthDocQA-Gov    & 0.9742 & 0.9667 & \multicolumn{3}{@{}l}{\textit{V3}} \\
SynthDocQA-Health.\ & 0.9963 & 0.9963 & Computer Science    & 0.7279 & 0.7209 \\
Tabfquad          & 0.8951 & 0.8957 & Energy               & 0.6682 & 0.6585 \\
Tatdqa            & 0.8141 & 0.8203 & Finance (EN)         & 0.5860 & 0.5791 \\
                  &        &        & Finance (FR)         & 0.4694 & 0.4505 \\
                  &        &        & HR                   & 0.5162 & 0.5498 \\
                  &        &        & Industrial           & 0.4699 & 0.4929 \\
                  &        &        & Pharmaceuticals      & 0.5958 & 0.5998 \\
                  &        &        & Physics              & 0.4782 & 0.4610 \\
\midrule
\textbf{V1 Average} & \textbf{0.9082} & \textbf{0.9105} & \textbf{V3 Average} & \textbf{0.5639} & \textbf{0.5641} \\
\bottomrule
\end{tabular}
\end{table}

\section{Per-Task Retrieval Results: \ours{}-0.8B}
\label{sec:scaling-pertask}

\begin{table}[H]
\centering
\caption{Per-task retrieval performance of \ours{}-0.8B. Source: \texttt{results/vidore/} in the public model repo.}
\label{tab:scaling-pertask}
\small
\begin{tabular}{@{}lc@{\hskip 1.5em}lc@{}}
\toprule
\multicolumn{2}{c}{\textit{ViDoRe V1 (10 tasks)}} & \multicolumn{2}{c}{\textit{ViDoRe V2 \& V3}} \\
\cmidrule(r){1-2} \cmidrule(l){3-4}
Task & \ndcg{} & Task & \ndcg{} \\
\midrule
ArxivQA           & 0.8511 & \multicolumn{2}{@{}l}{\textit{V2}} \\
DocVQA            & 0.6134 & BioMedical Lectures  & 0.5423 \\
InfoVQA           & 0.9061 & ESG Reports (HL)    & 0.5678 \\
ShiftProject      & 0.7682 & ESG Reports         & 0.4954 \\
SynthDocQA-AI     & 0.9819 & Economics Reports   & 0.5212 \\
SynthDocQA-Energy & 0.9628 & \textbf{V2 Average} & \textbf{0.5317} \\
SynthDocQA-Gov    & 0.9293 & \multicolumn{2}{@{}l}{\textit{V3}} \\
SynthDocQA-Health.\ & 0.9769 & Computer Science    & 0.6319 \\
Tabfquad          & 0.7848 & Energy               & 0.4908 \\
Tatdqa            & 0.7859 & Finance (EN)         & 0.4343 \\
                  &        & Finance (FR)         & 0.3044 \\
                  &        & HR                   & 0.4093 \\
                  &        & Industrial           & 0.3210 \\
                  &        & Pharmaceuticals      & 0.5508 \\
                  &        & Physics               & 0.4051 \\
\midrule
\textbf{V1 Average} & \textbf{0.8560} & \textbf{V3 Average} & \textbf{0.4434} \\
\bottomrule
\end{tabular}
\end{table}

\section{Per-Benchmark Generation Results}
\label{sec:generation-pertask}

Base Qwen3.5-4B via \texttt{model.generate()} vs.\ \ours{}-4B with LoRA disabled on four VQA benchmarks under greedy decoding ($T{=}0$, 128 new tokens). Exact Match\% is the fraction of base/\ours{} output pairs that are byte-identical strings.

\begin{table}[H]
\centering
\caption{Generation equivalence across four VQA benchmarks. Base: Qwen3.5-4B via \texttt{model.generate()}; \ours{}: same weights with LoRA disabled, using the custom KV-cache path. Greedy decoding, 128 new tokens. Max $|\Delta|{=}0.0043$ across all four benchmarks (DocVQA, the largest sample with the highest base ANLS, dominates the comparison).}
\label{tab:generation}
\small
\begin{tabular}{@{}lrccrr@{}}
\toprule
Benchmark & $n$ & Base ANLS & \ours{}-4B ANLS & $\Delta$ & Exact Match\% \\
\midrule
DocVQA    & 5{,}000 & 0.5449 & 0.5490 & $+$0.0041 & 71.08\% \\
ChartQA$^\dagger$   & 2{,}500 & 0.0015 & 0.0015 & $+$0.0000 & 22.92\% \\
InfoVQA   & 2{,}801 & 0.1804 & 0.1761 & $-$0.0043 & 31.06\% \\
TextVQA$^\dagger$   & 5{,}000 & 0.0566 & 0.0574 & $+$0.0008 & 16.20\% \\
\midrule
\textbf{Total} & \textbf{15{,}301} & & & & \\
\bottomrule
\end{tabular}

\smallskip
\textsuperscript{$\dagger$}\,On ChartQA and TextVQA the \emph{base} Qwen3.5-4B ANLS is at the noise floor (0.0015 and 0.0566 respectively) under our prompt format: Qwen3.5-4B is not instruction-tuned for these benchmarks' expected short-answer format, so its raw outputs do not match references at the surface-form level that ANLS scores. The \ours{} equivalence claim on these rows is the \emph{delta} ($\Delta$, second-to-last column), not the absolute score: both paths produce the same near-zero output. We report these benchmarks for completeness; rely on DocVQA and InfoVQA, where base ANLS is in a meaningful range, as the load-bearing equivalence evidence.
\end{table}

\section{Released Artifacts}
\label{sec:artifacts}

Table~\ref{tab:artifacts} consolidates the scripts and report JSONs referenced inline. All paths are relative to the corresponding HuggingFace model repository.

\begin{table}[H]
\centering
\caption{Released artifacts: script $\rightarrow$ claim $\rightarrow$ report JSON. Repos: \texttt{athrael-soju/HydraQwen3.5-4B}, \texttt{-0.8B}, \texttt{HydraQwen2.5-Omni-3B}.}
\label{tab:artifacts}
\small
\begin{tabular}{@{}llll@{}}
\toprule
Section & Script & Claim & Report \\
\midrule
\Cref{sec:requirements} (Req.\ 3) & \texttt{scripts/bench\_kv\_cache.py} & KV-cache speedup ($7.65\times$ mean) & \texttt{results/efficiency/kv\_cache\_bench/report.json} \\
\Cref{sec:generation} & \texttt{scripts/test\_gen\_equivalence\_4b.py} (4B), \texttt{scripts/test\_gen\_equivalence.py} (0.8B) & 426/426 bitwise match & \texttt{results/generation\_equivalence/} \\
\Cref{sec:experiments} (Eff.) & \texttt{scripts/bench\_mode\_switch\_vram\_4b.py} (4B), \texttt{scripts/bench\_mode\_switch\_vram.py} (0.8B) & Peak VRAM at both scales & \texttt{results/mode\_switch\_vram/} \\
\Cref{sec:kv-isolation} & \texttt{results/kv\_cache\_isolation/run.py} & 50/50 byte-identical & \texttt{results/kv\_cache\_isolation/kv\_isolation\_report.json} \\
\Cref{sec:omni} & (in Omni repo) & InfoVQA generation equivalence & \texttt{results/infovqa\_report.json} \\
\bottomrule
\end{tabular}
\end{table}

\section{Omni-Modal Per-Task Results}

\begin{table}[H]
\centering
\caption{\ours{}-Omni image retrieval on ViDoRe V1, V2, and V3. Model: \texttt{vidore/colqwen-omni-v0.1} (Qwen2.5-Omni-3B backbone, 4.4B total parameters). No \ours{}-specific training. V1 is the full 10-task set, directly comparable to the \ours{}-4B V1 column in \Cref{tab:vidore-pertask}.}
\label{tab:omni-vidore}
\small
\begin{tabular}{@{}lc@{\hskip 2em}lc@{\hskip 2em}lc@{}}
\toprule
\multicolumn{2}{c}{\textit{ViDoRe V1}} & \multicolumn{2}{c}{\textit{ViDoRe V2}} & \multicolumn{2}{c}{\textit{ViDoRe V3}} \\
\cmidrule(r){1-2} \cmidrule(lr){3-4} \cmidrule(l){5-6}
Task & \ndcg{} & Task & \ndcg{} & Task & \ndcg{} \\
\midrule
SynthDocQA-AI         & 0.9852 & ESG Reports (HL)    & 0.6050 & Computer Science & 0.6727 \\
SynthDocQA-Healthcare & 0.9663 & BioMedical Lectures & 0.5827 & Energy           & 0.5574 \\
SynthDocQA-Energy     & 0.9566 & ESG Reports         & 0.4860 & Pharmaceuticals  & 0.5446 \\
SynthDocQA-Gov        & 0.9529 & Economics Reports   & 0.4676 & HR               & 0.4953 \\
InfoVQA               & 0.9341 &                     &        & Finance (EN)     & 0.4636 \\
Tabfquad              & 0.8891 &                     &        & Physics          & 0.4180 \\
ArxivQA               & 0.8677 &                     &        & Industrial       & 0.4012 \\
ShiftProject          & 0.8398 &                     &        & Finance (FR)     & 0.3731 \\
Tatdqa                & 0.8196 &                     &        &                  &        \\
DocVQA                & 0.6537 &                     &        &                  &        \\
\midrule
\textbf{V1 Average}   & \textbf{0.8865} & \textbf{V2 Average} & \textbf{0.5353} & \textbf{V3 Average} & \textbf{0.4907} \\
\bottomrule
\end{tabular}
\end{table}

\begin{table}[H]
\centering
\caption{\ours{}-Omni generation equivalence on InfoVQA, using the same protocol as \Cref{tab:generation}. Base: Qwen2.5-Omni-3B thinker via \texttt{Qwen2\_5OmniThinkerForConditionalGeneration}; \ours{}-Omni: \texttt{vidore/colqwen-omni-v0.1} with LoRA disabled and \texttt{lm\_head} extracted from the thinker. Greedy decoding, 128 new tokens, with a short-answer prompt suffix applied identically to both paths (needed because Qwen2.5-Omni's default outputs are sentence-form). 95\% CI via 10{,}000-sample bootstrap.}
\label{tab:omni-generation}
\small
\begin{tabular}{@{}lrcccc@{}}
\toprule
Benchmark & $n$ & Base ANLS & \ours{}-Omni ANLS & $\Delta$ (95\% CI) & Exact Match\% \\
\midrule
InfoVQA & 2{,}801 & 0.7257 & 0.7257 & $+$0.0000 [$+$0.0000, $+$0.0000] & 100.00\% \\
\bottomrule
\end{tabular}
\end{table}

\bibliography{references}

\end{document}